\documentclass[letterpaper, 10 pt, conference]{ieeeconf}
\let\labelindent\relax
\usepackage{amsmath,amsfonts,amssymb,bm}
\usepackage{algorithm}
\usepackage{algpseudocode}
\usepackage{array}
\usepackage[caption=false,font=normalsize,labelfont=sf,textfont=sf]{subfig}
\usepackage{textcomp}
\usepackage{stfloats}
\usepackage{url}
\usepackage{verbatim}
\usepackage{graphicx}
\usepackage{subfig} 
\usepackage{caption} 
\usepackage{overpic}
\usepackage{cite}
\usepackage{pdflscape}
\usepackage{rotating}
\usepackage{makecell}
\usepackage[table]{xcolor}
\usepackage{hyperref}
\usepackage{listings}
\usepackage{color}
\usepackage{tikz}
\usepackage{enumitem}
\usepackage[normalem]{ulem}
\usetikzlibrary{arrows,positioning,calc}
\newcommand*{\SuperScriptSameStyle}[1]{%
  \ensuremath{%
    \mathchoice
      {{}^{\displaystyle #1}}%
      {{}^{\textstyle #1}}%
      {{}^{\scriptstyle #1}}%
      {{}^{\scriptscriptstyle #1}}%
  }%
}

\newcommand*{\oneS}{\SuperScriptSameStyle{*}}
\newcommand*{\twoS}{\SuperScriptSameStyle{**}}
\newcommand*{\threeS}{\SuperScriptSameStyle{*{*}*}}

\newenvironment{blueadd}
    {} 
    {} 
\newcommand{\reddelete}[1]{}

\DeclareMathOperator*{\argmin}{arg\,min}

\definecolor{turquoise}{cmyk}{0.65,0,0.1,0.1}
\definecolor{purple}{rgb}{0.65,0,0.65}
\definecolor{darkgreen}{rgb}{0.0, 0.5, 0.0}
\definecolor{darkred}{rgb}{0.5, 0.0, 0.0}
\definecolor{darkblue}{rgb}{0.0, 0.0, 0.5}
\definecolor{blue}{rgb}{0.0, 0.0, 1.0}
\definecolor{orange}{rgb}{1.0, 0.5, 0.0}
\definecolor{red}{rgb}{1.0, 0.0, 0.0}
\definecolor{cherry}{RGB}{186,12,47}

\newcommand{\refeq}[1]{Eq.~\ref{#1}}

\begin{document}

\title{AeroHaptix: A Wearable Vibrotactile Feedback System for \\Enhancing Collision Avoidance in UAV Teleoperation}

\author{Bingjian Huang, Zhecheng Wang, Qilong Cheng, Siyi Ren, Hanfeng Cai, \\Antonio Alvarez Valdivia, Karthik Mahadevan, and Daniel Wigdor}



\maketitle

\begin{abstract}



Haptic feedback enhances collision avoidance by providing directional obstacle information to operators during unmanned aerial vehicle (UAV) teleoperation. However, such feedback is often rendered via haptic joysticks, which are unfamiliar to UAV operators and limited to single-direction force feedback. Additionally, the direct coupling between the input device and the feedback method diminishes operators' sense of control and induces oscillatory movements. To overcome these limitations, we propose AeroHaptix, a wearable haptic feedback system that uses spatial vibrations to simultaneously communicate multiple obstacle directions to operators, without interfering with their input control. The layout of vibrotactile actuators was optimized via a perceptual study to eliminate perceptual biases and achieve uniform spatial coverage. A novel rendering algorithm, MultiCBF, extended control barrier functions to support multi-directional feedback. Our system evaluation showed that compared to a no-feedback condition, AeroHaptix effectively reduced the number of collisions and input disagreement. Furthermore, operators reported that AeroHaptix was more helpful than force feedback, with improved situational awareness and comparable workload.

\end{abstract} 


\section{Introduction}


\footnote{This work has been submitted to the IEEE for possible publication. Copyright may be transferred without notice, after which this version may no longer be accessible.}Unmanned aerial vehicle (UAV) teleoperation enables operators to pilot UAVs beyond their visual line of sight, enabling task completions in remote and hazardous environments. However, teleoperation is challenging due to the physical separation between the operator and the UAV, which limits the operator's ability to perceive obstacles and avoid collisions. To solve this problem, researchers have utilized haptic feedback and collision avoidance algorithms to convey obstacle information (e.g., parametric risk fields~\cite{lam2009artificial}, time-to-impact~\cite{brandt2010haptic}, dynamic kinesthetic boundary~\cite{hou2015dynamic}, and control barrier functions (CBF)~\cite{zhang2020haptic}).

Despite advancements in algorithms, the devices used to render haptic feedback are primarily haptic joysticks with three degree-of-freedom (DoF) force feedback~\cite{novint_falcon, the_touch, force_dimension}. While proven effective for collision avoidance~\cite{zhang2021haptic,philbrick2014effects}, they are rarely adopted for real-world operation due to the high cost of transitioning from standard radio control (RC) controllers. Their usage is also constrained to indoor settings as haptic joysticks must be surface-mounted to provide accurate feedback. Moreover, their limited information bandwidth confines force feedback to a single direction~\cite{brandt2009haptic,zhang2020haptic}, compromising situational awareness in multi-obstacle environments. Since haptic joysticks exert force feedback on the hand, the direct coupling of input and output channels impairs control precision, causing oscillatory behavior~\cite{philbrick2014effects} and low user acceptance~\cite{ho2018increasing}. \reddelete{Alternative devices such as cable-driven exoskeletons [12],[13] solved the mobility issue, but added new challenges such as physical fatigue and limited degrees of freedom.}\begin{blueadd}Alternative devices such as cable-driven exoskeletons~\cite{rognon2018haptic,rognon2019haptic,ramachandran2022arm} address mobility constraints but have limited degrees of freedom and introduce physical fatigue. Wearable devices like haptic gloves~\cite{macchini2020hand} and waistbands~\cite{spiss2018comparison} permit intuitive hand-motion control, but have limited numbers of actuators and bandwidth.\end{blueadd}

\begin{figure}
    \centering
    \includegraphics[width=\linewidth]{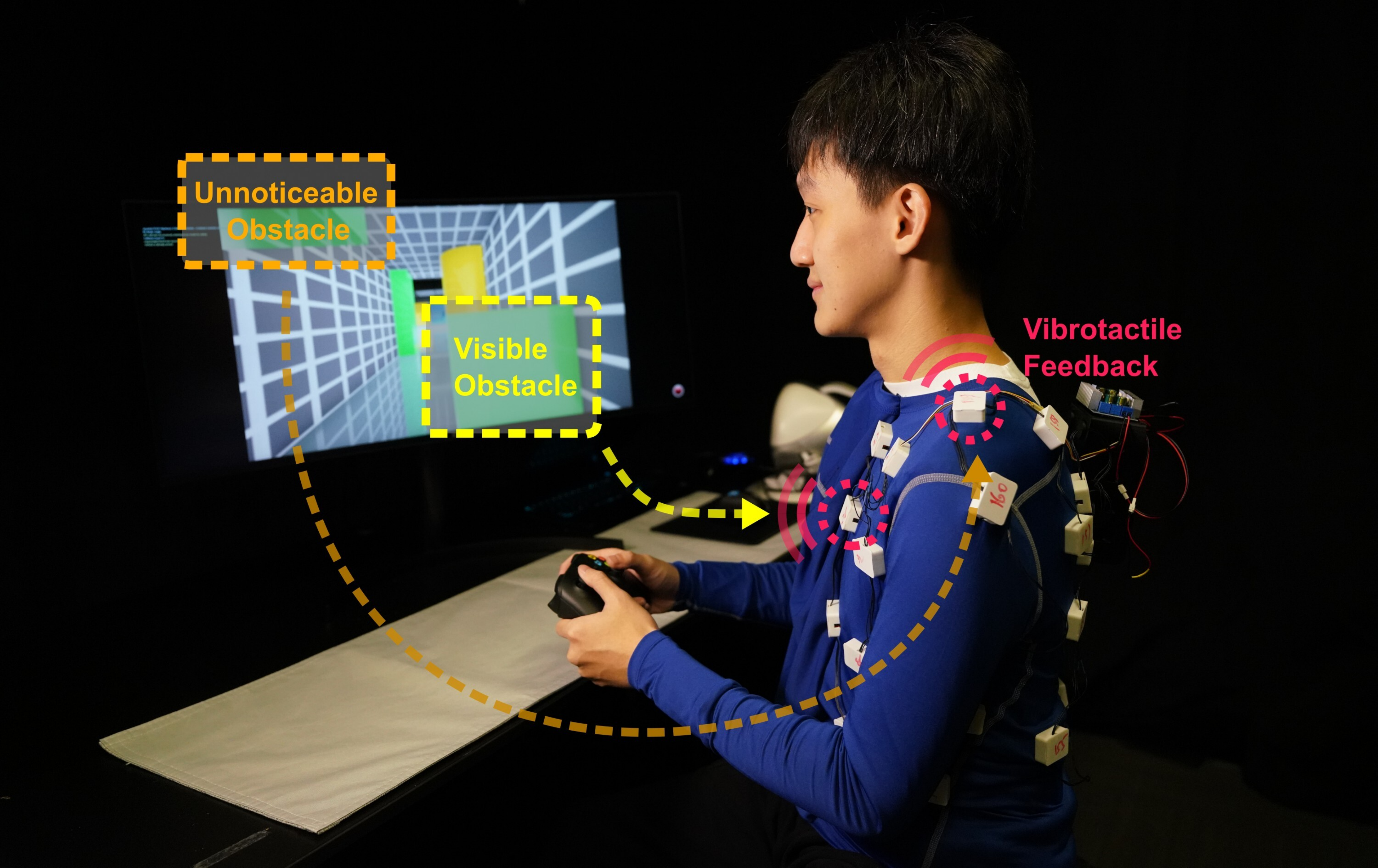}
    \caption{AeroHaptix assists UAV operators with collision avoidance by delivering on-body spatial vibrotactile feedback about obstacle directions. Operators not only see and feel visible obstacles (yellow), but also perceive obstacles out of view (orange).}
    \label{fig:sec5_experiment_scene}
    \vspace{-10pt}
\end{figure}

To address these limitations, we designed a wearable haptic feedback system, AeroHaptix, that renders vibrotactile feedback at 32 upper body positions to convey obstacle directions. Since vibrations are delivered to the body, they do not hinder hand movement and can be seamlessly integrated with existing teleoperation workflows (i.e., RC controllers). We conducted a perceptual study with ten participants to map body positions to spatial directions and employed a neural network to optimize the layout of the vibrotactile actuators. We also designed \emph{MultiCBF}, a novel collision avoidance algorithm that extends CBF to render multi-directional haptic feedback. Our evaluation showed that AeroHaptix effectively reduced collisions and input disagreements relative to a no-feedback condition. Participants also rated AeroHaptix more helpful than the force feedback method, with improved situational awareness and comparable workload.




\section{Related Work}




Of most relevance to the present research is prior literature on collision avoidance algorithms for UAV teleoperation and vibrotactile feedback to convey spatial information.

\subsection{Collision Avoidance Algorithms}

Collision avoidance algorithms were first used for unmanned ground vehicle teleoperation. In 1998, Hong et al. proposed an artificial force field to compute virtual forces from the potential fields of vehicles and obstacles ~\cite{hong1998artificial}. Later, Boschloo et al. and Lam et al. adopted this concept for UAVs and developed basic risk field~\cite{boschloo2004collision} and parametric risk field~\cite{lam2009artificial}. These methods reduced collisions but had difficulties navigating through narrow corridors. Brandt and Colton introduced time-to-impact and virtual string algorithms to compute virtual impact forces between a UAV and obstacles \cite{brandt2010haptic}. These algorithms reduced collisions and workloads but caused oscillatory movements in cluttered environments \cite{philbrick2014effects}. 

Alternatively, some researchers proposed algorithms that overrode user input with safer commands, such as dynamic kinesthetic boundary~\cite{hou2013dynamic,hou2015dynamic} and obstacle avoidance system for teleoperation~\cite{courtois2022oast}. While these algorithms eliminated collisions, they reduced operators' sense of control and user acceptance. More recently, Zhang et al. proposed using control barrier functions (CBF)~\cite{zhang2020haptic}, which modified an input control signal to closely match the original while adhering to safety constraints. They also designed a haptic shared autonomy control scheme that enhanced operators' sense of control~\cite{zhang2021haptic}. We extended their CBF algorithm to convey haptic feedback for multiple obstacles simultaneously.

\subsection{Vibrotactile Feedback to Convey Spatial Information} 

Vibrotactile feedback conveys sensory information to humans via actuators that vibrate on the skin. Its utility has been demonstrated in various scenarios where spatial awareness is crucial, including headbands for locating 3D objects in virtual reality~\cite{louison2017spatialized, de2017designing} and wearable actuators aiding obstacle detection for blind and visually impaired users~\cite{lee2021goldeye, flores2015vibrotactile, wang2017enabling}. In robotics, vibrations have been used to indicate handover positions and robotic arm trajectories~\cite{zaffir2024presentation, grushko2021intuitive}, hydrodynamic flow near underwater robots~\cite{xia2022virtual}, and obstacle positions around UGVs~\cite{de2011enhancing}.

However, delivering obstacle directions during UAV teleoperation is challenging because it requires numerous actuators to precisely represent obstacle directions. Existing vibrotactile systems like TactJam~\cite{wittchen2022tactjam} and VHP~\cite{dementyev2021vhp}) offer customizability but only support up to twelve actuators. In contrast, bHaptics X40~\cite{bHapticsWebsite}) included more actuators but lacked flexible placement and body coverage. To address these limitations, we developed custom hardware to support high-density actuators with increased customizability.

\section{AeroHaptix Hardware Design} \label{section3:hardware}

To assist UAV teleoperation, AeroHaptix should:
\begin{itemize}
    \item \textbf{R1}: support the fine-grained control of numerous actuators to distinguish obstacles from different directions;
    \item \textbf{R2}: support the reconfiguration of actuator layouts so actuators could be adjusted as needed; and
    \item \textbf{R3}: \reddelete{support low-latency communication so multi-point feedback could be activated without a noticeable delay. }\begin{blueadd}supports low-latency data communication (i.e.,  an end-to-end latency of less than 45 ms~\cite{vogels2004detection} and a chain communication latency of less than 20 ms \cite{hirsh1961perceived} to ensure synchronized multi-point feedback).\end{blueadd}
\end{itemize}




\begin{figure}[h]
    \centering
    \includegraphics[width=0.9\linewidth]{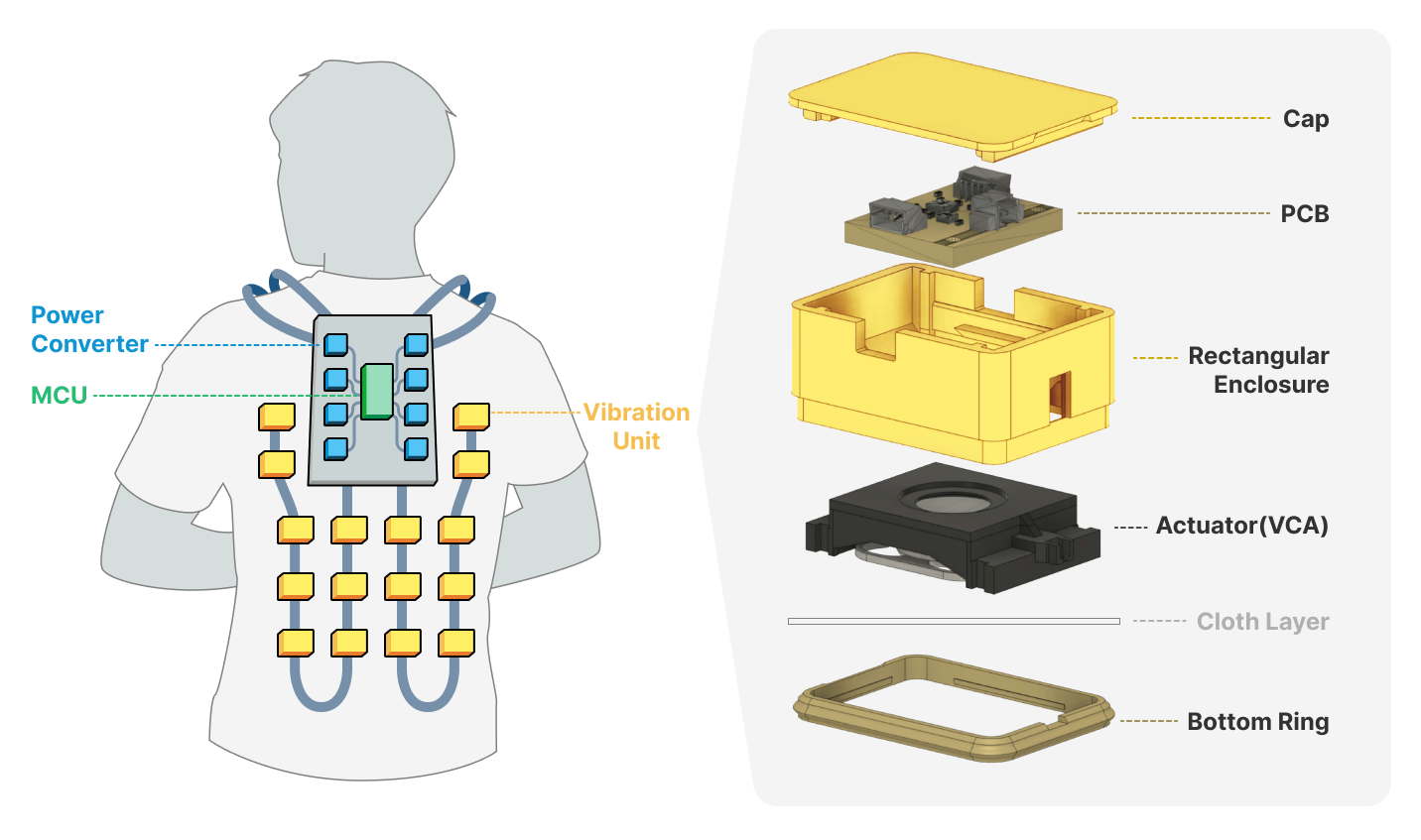}
    \caption{Aerohaptix's hardware design, with an exploded view of a vibration unit.}
    \label{fig:sec3_unit_design}
    \vspace{-10pt}
\end{figure}

AeroHaptix has a central unit and multiple chains of vibration units (Figure \ref{fig:sec3_unit_design}). Each vibration unit contains a 32~\(\times\)~22~mm voice coil actuator (PUI Audio HD-VA3222), a custom PCB, and 3D-printed parts. The actuator covers frequencies from 80--500\,Hz and reaches a peak acceleration of 2.52\,Gp-p at 133\,Hz with 1.5\,\(V_{\mathrm{rms}}\). The PCB features a PIC16F18313 MCU and a DRV8837 H-bridge motor driver. It receives and transmits commands via UART and generates waveforms. Upon receiving a \textit{start} command, the MCU drives the actuator with fine-grained control (\textbf{R1}) over 16 intensity levels and 8 frequencies. The 3D-printed parts include a cap and an enclosure for stability, and a bottom ring that \reddelete{enabled easy repositioning
on garment (R2) via press-fit.}\begin{blueadd}attaches vibration units to garments (\textbf{R2}) via pressed fit. The garment used herein is an off-the-shelf compression shirt, but the system can be easily adapted to other garments.\end{blueadd}

\begin{figure}[h]
    \centering
    \includegraphics[width=0.9\linewidth]{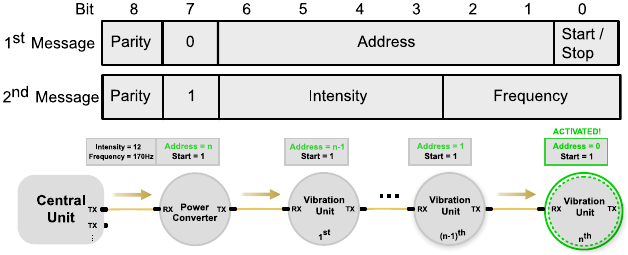}
    \caption{The data transmission on each chain used a two-byte UART protocol. The central unit sent a command with address n, and each unit deducted the address by 1 until it reached the target unit.}
    \label{fig:sec3_protocol}
    \vspace{-10pt}
\end{figure}

To support low-latency feedback (\textbf{R3}), we designed a chain-connection topology and a custom \begin{blueadd}universal asynchronous receiver-transmitter\end{blueadd} (UART) protocol (Figure \ref{fig:sec3_protocol}). Each chain has a central unit, a power converter, and support up to 20 vibration units. \begin{blueadd}The central unit receives Bluetooth commands from a PC, converts them to UART commands, and sends them along the chain.\end{blueadd} Each vibration unit examines the incoming UART commands to determine whether to execute them or forward them to the next unit. The UART protocol transmits two-byte messages at a baud rate of 115.2 kHz. The first byte contains the target unit address and a \textit{start/stop} bit. The second byte contains vibration parameters (i.e., intensity and frequency). A technical evaluation found \reddelete{there was a 125 $\mu$s delay between two vibration units. With 20 units, the total delay was 2.5 ms, fulfilling the requirement.}\begin{blueadd}5 ms of controller input latency, 1 ms of system processing latency, 14 ms of Bluetooth latency, and 2.5 ms of chain communication latency. Thus, the end-to-end latency was 23 ms, which was under the human perception threshold (\textbf{R3}).\end{blueadd}


\begin{blueadd}
AeroHaptix was designed with ergonomics in mind. The garment, central unit and vibration units weigh 1.5~kg, which is similar to a winter coat. The central unit, being the heaviest component, is mounted on the upper back to balance the weight and ensure a comfortable wearing experience.
AeroHaptix is also cost-effective. The central unit costs \$100 USD (i.e., \$19.95 for the MCU, \$20.10 for the PCB, \$52.75 for the LiPo batteries, and \$7.20 for the 3D printed parts). The vibration unit costs \$21.77 USD (i.e., \$3.29 for the PCB, \$18.09 for the actuator, and \$0.39 for the 3D printed parts). The final design comprising one central unit and 32 vibration units costs \$796.64 USD, which falls within the range of most haptic joysticks.
\end{blueadd}

\section{Haptic Actuator Layout Optimization}
\label{sec:4_determine_layout}

To achieve comprehensive obstacle direction coverage (\textbf{R1}), one naive approach is to uniformly distribute actuators around the body. However, previous research has shown that due to proprioceptive biases, a uniform actuator layout around the waist could lead to uneven coverage of 2D directions on the azimuth plane ~\cite{van2005presenting}. Therefore, we designed a perceptual study to collect data about position-direction mappings on the upper body and trained a neural network to generate an optimized actuator layout using this data.

Ten participants (6 male, 4 female; mean age = 25 years, std = 2 years) were recruited for the study. The study lasted 40 minutes and each participant received \$20 CAD as compensation. The study was approved by our university's Research Ethics Board.

\subsection{Data Collection Procedure} 

The initial actuator layout used in the study was a uniformly distributed grid of 46 actuators distributed on the upper body (i.e., red dots in Figure \ref{fig:final_layout}). Each actuator was positioned equidistantly from its neighbors, with $18$ actuators on the front and back, $3$ on each side of the waist, and $2$ on each shoulder. Following VibroMap~\cite{elsayed2020vibromap}, the inter-actuator distances were 8 cm.

\begin{figure}
    \centering
    \includegraphics[width=1.0\linewidth]{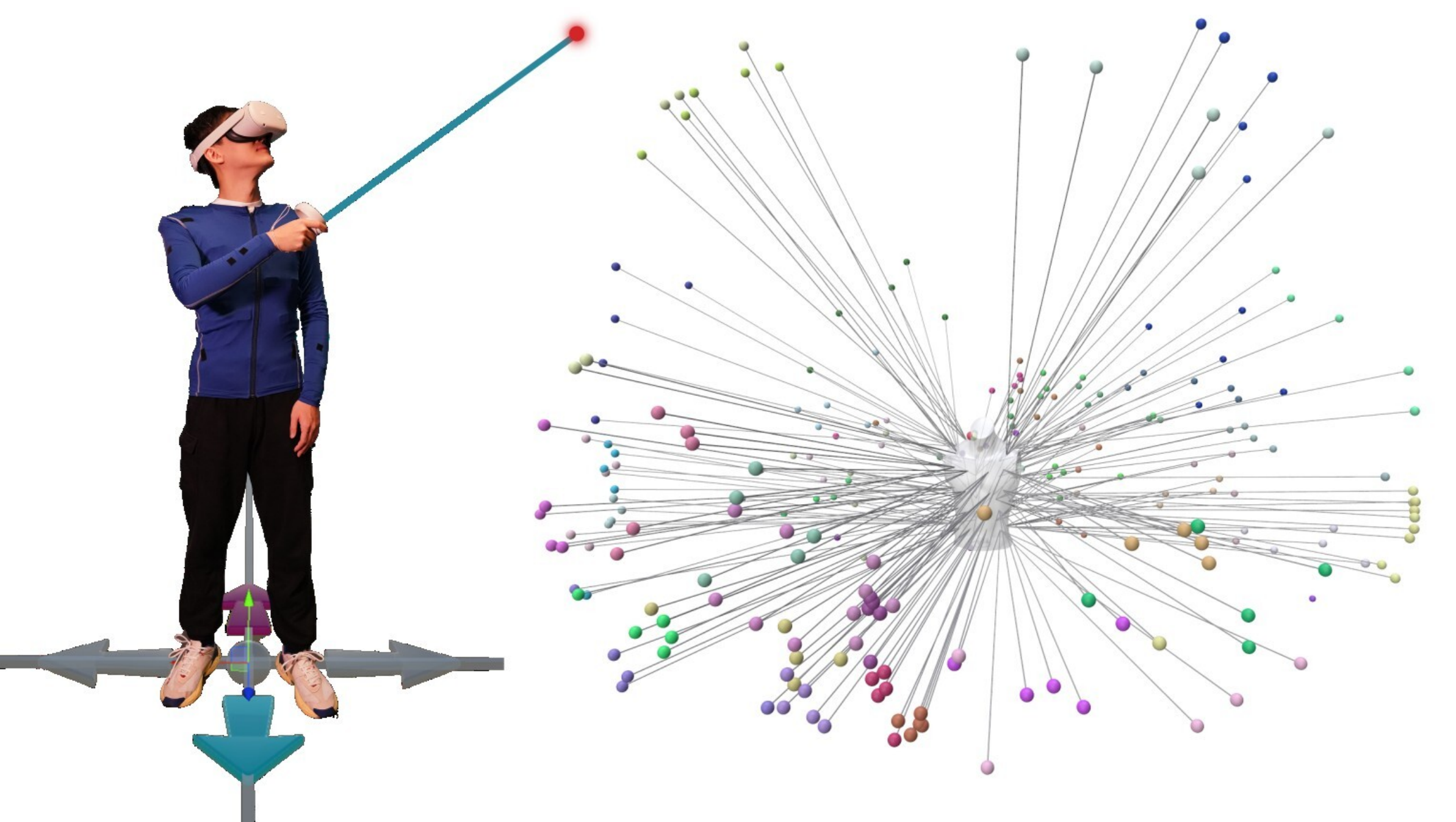}
    \caption{Left: In the virtual reality environment, participants pointed in a spatial direction after perceiving a vibrotactile cue. Right: Example data points collected from one participant, where points of the same color were from the same actuator.}
    \label{fig:experiment_scene}
    \vspace{-10pt}
\end{figure}

A Unity 3D application was developed for the study. Participants wore a Meta Quest~2 VR headset and stood in a virtual space with coordinates surrounding them (Figure \ref{fig:experiment_scene}). They received a vibrotactile cue from an actuator, pointed in the perceived obstacle direction using a Quest~2 controller, then clicked the trigger button to confirm the direction. Each participant completed 230 trials in a randomized order (i.e., 46 actuators x 5 repetitions).

\subsection{Layout Optimization}
We collected 2,300 mappings between the actuators \begin{blueadd}positioned\end{blueadd} on participants' body surfaces $p \in \Omega \subset \mathbb{R}^3$ and the obstacle directions participants reported $(\vartheta, \varphi)$ on a $2$-sphere $S^2$. \begin{blueadd}The average standard deviations of the perceived directions for each actuator were $\overline{\mathrm{SD}}_\vartheta = 0.18 \, \mathrm{rad}$ and $\overline{\mathrm{SD}}_\varphi = 0.53 \, \mathrm{rad}$. This suggested that a universal layout could be used across participants.\end{blueadd} The data was used to approximate an inverse mapping from $S^2$ to $\Omega \subset \mathbb{R}^3$ using a Multi-layer Perceptron neural network:
\begin{equation}
\label{eq:neural_haptics_mapping}
f_\theta = \argmin_{\theta} \sum_{i=1}^{N} \left\| f_\theta(\vartheta, \varphi) - p \right\|^2 + \lambda \sum_{i=1}^{N} \left | \nabla f_\theta(\vartheta, \varphi)\right |,
\end{equation}
where $\theta$ represented the neural network parameters and $\lambda$ was the regularization term weight. The network featured five $64$-unit hidden layers with \texttt{ReLU} activation. To promote mapping smoothness, we incorporated a total variation regularization term into the mean squared error loss function.


\subsection{Optimized Actuator Layout}
\label{sec:4_final_layout}

Previous studies have found that humans can discriminate force feedback directions on the hand with a $30^\circ$ resolution \cite{van2013anisotropy,tan2006force}.
To achieve similar discriminability on the body, we uniformly sampled 32 directions $\boldsymbol{\phi} = (\vartheta, \varphi)$ in 3D space $S^2$, where $\vartheta \in \{\frac{\pi}{6}, \frac{\pi}{3}, \frac{\pi}{2}, \frac{2\pi}{3}\}$ and $\varphi \in \{-\frac{3\pi}{4}, -\frac{\pi}{2}, -\frac{\pi}{4}, 0, \frac{\pi}{4}, \frac{\pi}{2}, \frac{3\pi}{4}, \pi\}$. Directions with $\vartheta = \frac{5\pi}{6}$ were omitted because they did not align with the upper body vibrations perceived by participants. 

\begin{figure}[h]
    \centering
    \begin{overpic}[width=0.9\linewidth]{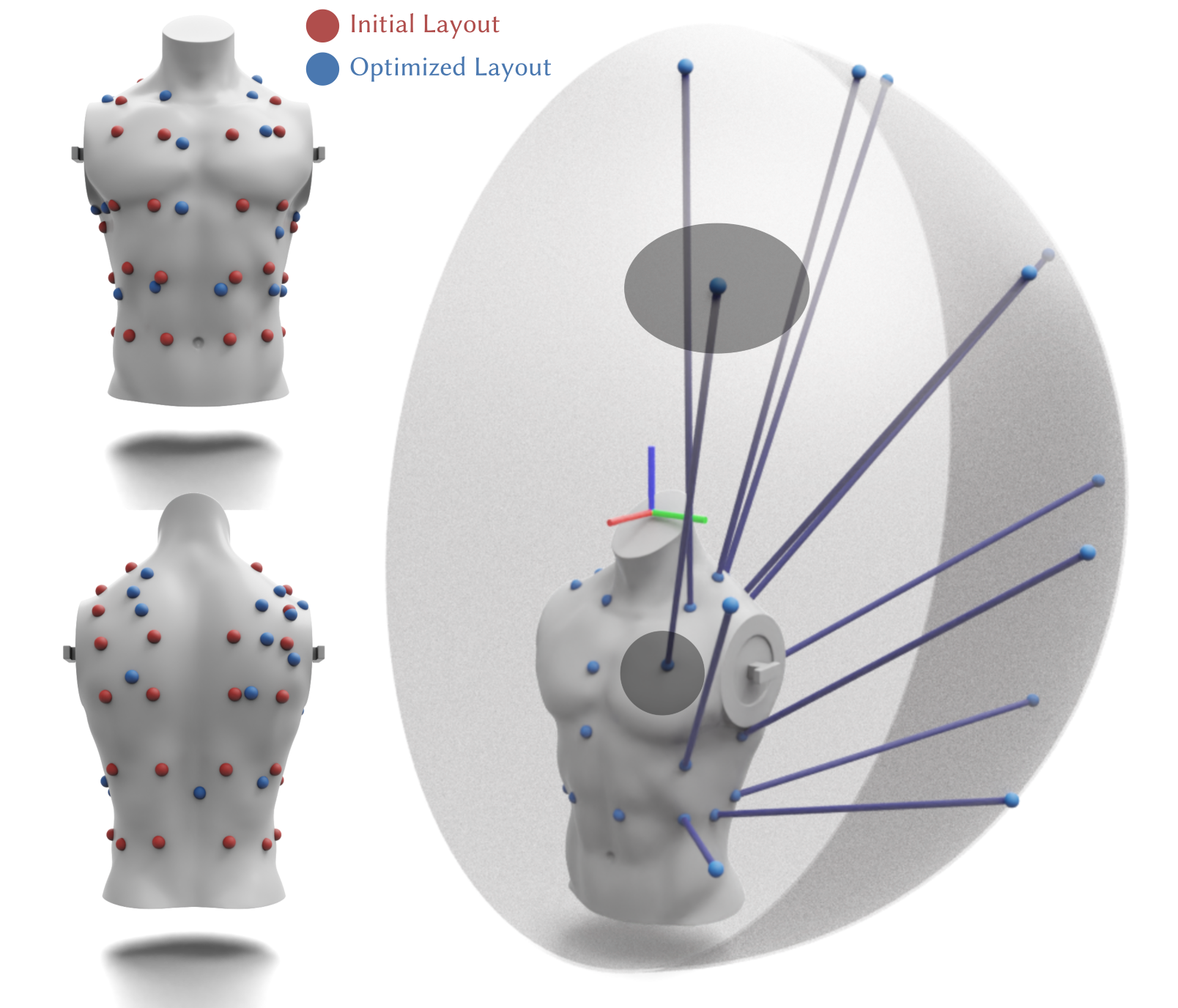}
        \put(56,60){\color{white}\small $(\vartheta, \varphi)$}
        \put(55,28){\color{white}\small $p$}
    \end{overpic}
    \caption{Red dots depict the initial uniformly distributed layout, while blue dots depict the  optimized layout. The lines represent the mappings between body positions and spatial directions on the left side of the body.}
    \label{fig:final_layout}
    \vspace{-10pt}
\end{figure}

We then used $f_\theta$ to find the corresponding actuator positions (i.e., blue dots in Figure \ref{fig:final_layout}). This optimized layout had more actuators positioned on body surfaces with greater curvatures (e.g., shoulders, waist). \begin{blueadd}
During the real-life setup, each actuator was localized relative to anatomical landmarks that were visible on the mannequin and the virtual model, such as collarbones, spine midlines and hip ridges. The overall setup time was around 30 minutes.\end{blueadd}

In our collision avoidance algorithm, MultiCBF, the optimized layout was represented by a set of 32 actuators $\mathcal{A} = \{\mathcal{A}_i\}_{i=1}^{32}$ and the corresponding direction set $\hat{r}$, where $\hat{r}_i$ was the direction associated with actuator $\mathcal{A}_i$.

\section{Collision Avoidance Algorithm}
\label{sec5:algorithm}

Existing collision avoidance algorithms such as PRF~\cite{lam2007collision} and CBF~\cite{zhang2020haptic} only support single-directional feedback due to the limited bandwidth of haptic joysticks. Since AeroHaptix supports spatial vibrations (\textbf{R3}), We extended CBF \cite{zhang2020haptic} to build a multi-directional feedback algorithm, \emph{MultiCBF}:

\begin{enumerate}[leftmargin=3.5mm]
\label{alg:multicbf}
    \item
        Considering that the continuous-time dynamics of a UAV can be modeled with a double integrator, where the control input $u$ corresponds to the acceleration command of the UAV. Let $x = \begin{bmatrix} q & \dot{q} \end{bmatrix}^T$ with $q$ and $\dot{q}$ being the position and the velocity of the UAV respectively. The dynamics of the system then become
        \begin{equation}\label{eqn:modified_double_integrator}
            \dot{x} = \begin{bmatrix} \dot{q} \\ \ddot{q} \end{bmatrix} = f(x) + g(x)u,
        \end{equation}
    where $x \in \mathcal{X} \subset \mathbb{R}^n$, $u \in \mathcal{U} \subset \mathbb{R}^m$, $f: \mathcal{X} \to \mathbb{R}^n$ and $g: \mathcal{X} \to \mathbb{R}^{n \times m}$ are Lipschitz continuous functions.
    
    \item 
        Consider a space with a set of obstacles $\mathcal{B} = \{b_i\}_{i=1}^{|\mathcal{B}|}$, where every obstacle $b_i$ is associated with the center of mass $q_{b_i}$. For each obstacle $b_i$, we construct a set of safety constraints $\mathcal{C}_i \in \mathcal{X}$ defined as
        \begin{equation}\label{eqn:safety_set}
            \mathcal{C}_i := \{x \in \mathcal{X} : h_i(x) \geq 0 \},
        \end{equation}
        where $h_i: \mathcal{X} \to \mathbb{R}$ is a continuously differentiable function that defines the safety boundary of obstacle $b_i$, $h_i(x) = 0 \Leftrightarrow x \in \partial \mathcal{C}_i$. \begin{blueadd}In practice, one might define 
        \[
        h_i(x) = \bigl\| q - q_{b_i} \bigr\| - d_{\min},
        \]
        where $d_{\min}>0$ is the minimal allowable distance for obstacle avoidance. The condition $h_i(x) \ge 0$ ensures that the UAV is outside the obstacle's unsafe region.\end{blueadd}

    \item We define the CBF $\mathcal{U}_{\texttt{CBF}}$ for second-order systems as the set of all control inputs that keep the systems in the safe set $C$. For the CBF associated with $C_i$
    \begin{equation}\label{eqn:cbf}
        \begin{split}
            \mathcal{U}_{\texttt{CBF}, i}(h_i(x)) = \left\{ u \in \mathcal{U} : \right. 
            L_f^2 h_i(x) + L_g L_f h_i(x) u \\ + 
            K [h_i(x)\text{ }L_f h_i(x)]^T \left. \right\} \geq 0,
        \end{split}
    \end{equation}
    Here, 
    \[
        L_f h_i(x) = \nabla h_i(x)^\top f(x), \quad 
        L_g h_i(x) = \nabla h_i(x)^\top g(x),
    \]
    denote the first-order Lie derivatives, while $L_f^2 h_i(x)$ is the second-order Lie derivative. $K$ is a positive constant that adjusts the safety margin.
    
    \item 
    Given user input $u_{\texttt{ref}} \in \mathcal{U}$, the optimization of a safe input $u_{\texttt{safe}, i}$ for obstacle $b_i$ and its $h_i(x)$ can be formulated as a quadratic program to find a \textbf{local safe input} $u_{\texttt{safe}, i} \in \mathcal{U}_{\texttt{CBF}, i}$ (\refeq{eqn:cbf}) that is closest to $u_{\texttt{ref}}$ 
    \begin{equation}\label{eqn:qp}
        u_{\texttt{safe}, i} = \underset{u \in \mathcal{U}}{\argmin} \ \frac{1}{2} \| u - u_{\texttt{ref}} \|^2 \text{ s.t. } u \in \mathcal{U}_{\texttt{CBF},i}(h_i(x)).
    \end{equation}

    If $ u_{\texttt{safe}, i} \neq u_{\texttt{ref}}$, it means the user input $u_{\texttt{ref}}$ violates the safety constraint $C_i$ and an actuator $\mathcal{A}_j$ is triggered to notify the user. The choice of actuator $\mathcal{A}_j$ is determined by the most aligned actuator direction $\hat{r}_k$
    \begin{equation}\label{eqn:actuator_projection}
            j = \underset{k \in \{1,\dots,32\}}{\arg\max}  \left(\frac{q_{b_i}-q}{\|q_{b_i}-q\|} \cdot \hat{r}_k\right).
    \end{equation}
    The vibration intensity $I_j$ of actuator $\mathcal{A}_j$ is then determined by the difference between the user input $u_{\texttt{ref}}$ and the safe input $u_{\texttt{safe},i}$ multiplied by a gain factor $K_v$
    \begin{equation}\label{eqn:intensity}
        I_j = \|K_v (u_{\texttt{safe},i} - u_{\texttt{ref}})\|.
    \end{equation}

\end{enumerate}


Previous CBFs only considered a global safe input $u_{\texttt{safe}}$ computed from a global safety set $h(x) = \{h_i(x)\}_{i=1}^{|\mathcal{B}|}$ \cite{zhang2020haptic}, which was not sufficient to represent multiple obstacles and hindered operator's situational awareness. In contrast, MultiCBF computed local safe input for each obstacle and rendered multi-point haptic feedback simultaneously.

\section{System Evaluation}
To evaluate AeroHaptix's ability to assist UAV teleoperation, we designed a study in which participants maneuvered a simulated quadrotor with a front-facing camera through cluttered tunnels. Twelve participants were recruited for the study (11 male, 1 female; mean age = 24 years, std = 3 years), with seven having previous experience operating UAVs. Each experiment lasted 70 minutes and each participant received \$40 CAD as compensation. The study was approved by our university's Research Ethics Board.

\subsection{Experimental Conditions and Hypotheses}

Participants experienced three \textbf{feedback conditions}: no feedback (\textbf{NA}), force shared control (\textbf{FSC}), and vibrotactile shared control (\textbf{VSC}). For the NA and VSC conditions, participants used an Xbox controller for input, which had a control mechanism similar to an RC controller. The output device was AeroHaptix with vibrotactile feedback rendered using MultiCBF. For the FSC condition, a Novint Falcon haptic joystick \cite{novint_falcon} was used for input and output, similar to previous studies \cite{omari2013bilateral, reyes2015outdoor}. Force feedback during FSC was rendered using baseline CBF \cite{zhang2020haptic}. \begin{blueadd}We did not implement the baseline CBF on vibrotactile feedback because AeroHaptix was designed as a highly integrated system, with MultiCBF being a core component that delivered multiple cues. Switching to the baseline CBF would change the semantic meaning of the feedback and potentially cause confusion.\end{blueadd}

\begin{figure}[h]
    \centering
    \includegraphics[width=\linewidth]{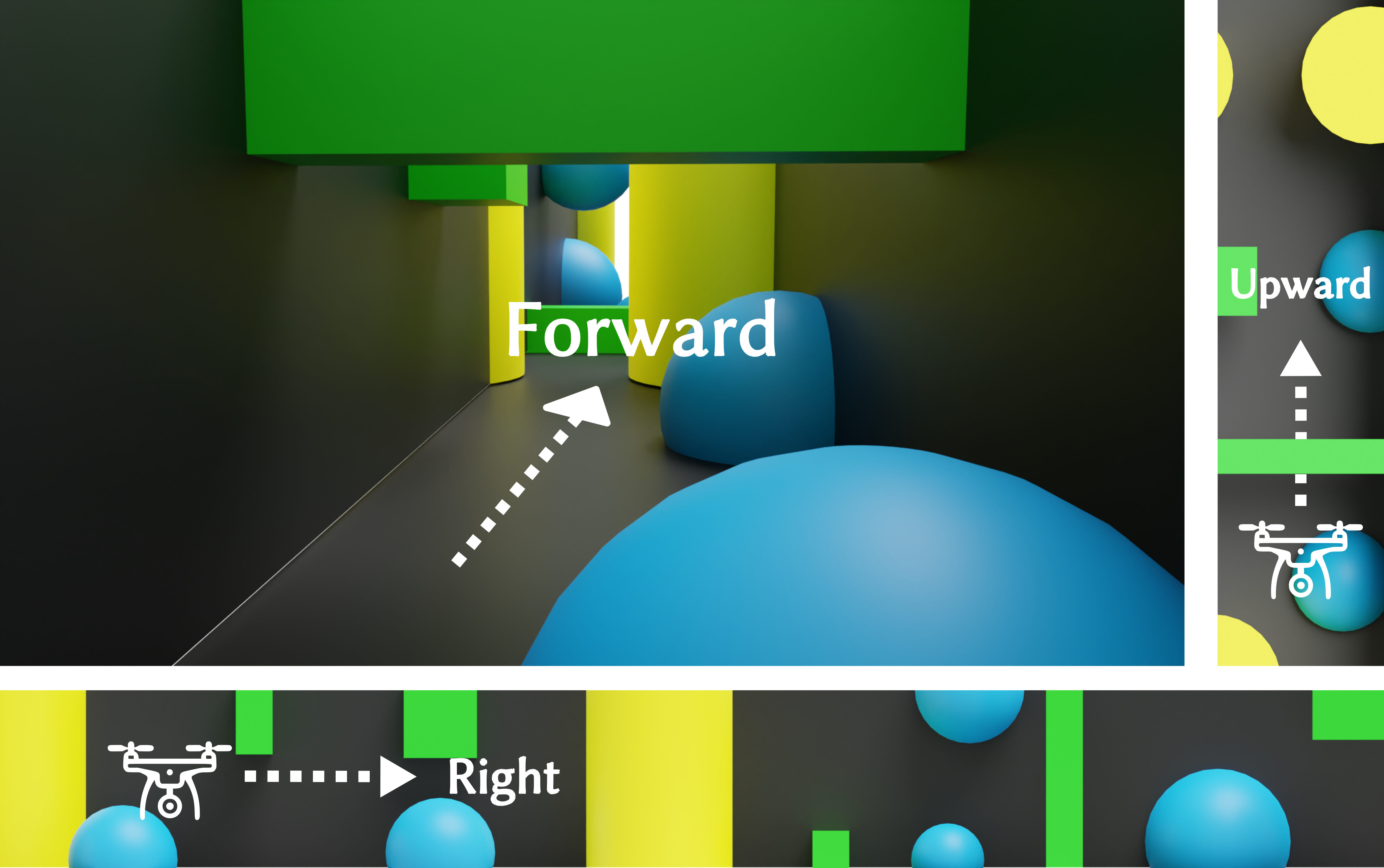}
    \caption{\begin{blueadd}The three flying directions used in the study: forward, right, and upward. Obstacles were randomly placed in the tunnel.\end{blueadd}}
    \label{fig:sec5_map}
\end{figure}

Because previous studies used complex experiment environments with varying visual capacities~\cite{courtois2022oast,hou2015dynamic, philbrick2014effects}, it was unclear when haptic feedback was beneficial. Thus, we used three \textbf{flying directions} \begin{blueadd}(Figure \ref{fig:sec5_map})\end{blueadd} to assess how visual capacity influenced participants' reliance on haptic feedback: forward (\textbf{FWD}), right (\textbf{R}) and upward (\textbf{UP}). As visual capacity was limited in right and upward directions, this may increase the difficulty for collision avoidance and illustrate the benefit of haptic feedback.

Our hypotheses were that:
\begin{enumerate}[label=(H$_{\arabic*}$)]
    \item compared to NA, VSC would reduce the number of collisions and input disagreement
    \item compared to flying forward, the number of collisions would be higher when flying right and upward
    \item compared to FSC, VSC would reduce task workload
    \item compared to FSC, VSC would increase participants' sense of control
    \item compared to FSC, VSC would be more effective at increasing situational awareness
\end{enumerate}

\subsection{Experimental Setup and Procedure}
The experimental scene was a $5\times5\times50$m tunnel (Figure~ \ref{fig:sec5_map}) that could face forward, right, or upward depending on the flying direction. The scene contained four planes and fifteen obstacles that were cubes, spheres, or cylinders. The safety boundary of plane $b_i$ was defined as: $h_i(x) = h_i(\begin{bmatrix}q & \dot{q}\end{bmatrix}^T) = (q-q_{b_i}) \times \hat{n}_{b_i}$, where $q_{b_i}$ was the center of mass and $\hat{n}_{b_i}$ was the unit normal of plane $b_i$. The safety boundaries of other obstacles were approximated using superellipsoids $h_i(x) = h(\begin{bmatrix}q & \dot{q}\end{bmatrix}^T) = (\frac{q_1-q_{b_i,1}}{a_1})^n + (\frac{q_2-q_{b_i,2}}{a_2})^n + (\frac{q_3-q_{b_i,3}}{a_3})^n$, where $q_{b_i}$ was the center of mass and $a$ was the scaling vector of obstacle $b_i$. Obstacle positions were randomized to avoid learning effects.

The study was performed in a Microsoft AirSim \cite{airsim2017fsr} simulated environment on an Alienware M15 Laptop with RTX 3060 GPU. \begin{blueadd}The baseline CBF and MultiCBF algorithms were implemented as Python custom libraries and ran on a Python server. On each simulation frame, the server detected input commands from the controller, received UAV state updates from AirSim, calculated collision risks using CBF algorithms, and sent vibration or force feedback commands to the corresponding output device if haptic feedback was enabled. At the end of each frame, the server updated the UAV's state in AirSim.\end{blueadd}

At the beginning of the study, the participant reviewed and signed a consent form. The participant was then briefed on the purpose and procedure of the study. Following this, the participant experienced the three feedback conditions in a randomized order ($3\times3$ Latin square). For each feedback condition, the participant underwent a practice round using the designated devices for five minutes. Then, they maneuvered the UAV to fly through three tunnels, one for each flying direction. After each feedback condition, the participant completed a questionnaire about task workload and their sense of control.

\subsection{Metrics and Analysis}

We collected objective and subjective measurements. Objective measurements included the total distance travelled, number of collisions, and input disagreement (computed as \reddelete{the difference between participant input and safe input }$\|u_{\texttt{ref}} - u_{\texttt{safe}}\|$), which evaluated the collision avoidance performance (\textbf{H1,H2}). For each metric, we conducted a two-way repeated measures ANOVA using SPSS. If significant effects were found, post-hoc pairwise comparisons using a Bonferroni correction were conducted.

Subjective measurements included NASA-TLX \cite{hart1988development} questions that evaluated task load (\textbf{H3}), and four 7-point Likert questions adapted from \cite{zhang2021haptic} that probed sense of control (\textbf{H4}):
\begin{enumerate}
    \item How easy was it to control drone with this input device?
    \item How much control did you feel you had over the drone?
    \item How well did the drone's motion match your intention?
    \item If you felt haptic feedback, how much did the feedback help you navigate the robot?
\end{enumerate}
\reddelete{For each metric, we conducted a Friedman test. If significant effects were found, post-hoc pairwise comparisons using Wilcoxon signed-rank tests were conducted.}\begin{blueadd}Friedman tests were conducted for these metrics, with significant effects analyzed via Wilcoxon signed-rank tests.\end{blueadd}

\reddelete{To evaluate situational awareness (H4), a pop-up window appeared at a random time during the trial for each direction (with the simulation paused) and participants were asked to report any perceived obstacles. This technique was adapted from situational awareness probing techniques like SAGAT and SPAM. Reported obstacles were categorized as visual obstacles that were visible at pause and haptic obstacles that were not seen but perceived through haptic feedback. As the visual capacity was consistent across conditions, haptic obstacles represented the enhanced situational awareness through haptic feedback. A two-way repeated measures ANOVA was run to determine the effect of the haptic feedback conditions and flying directions on situational awareness.}

\begin{blueadd}
To assess situational awareness (\textbf{H5}), the simulation was randomly paused and a pop-up asked participants to report any perceived obstacles. This technique was adapted from situational awareness probing techniques, e.g., SAGAT and SPAM \cite{endsley2021systematic}. Reported obstacles were categorized as \textit{visual obstacles} that were visible at pause and \textit{haptic obstacles} that were perceived through haptic feedback. A two-way repeated measures ANOVA was used to analyze the effects of haptic feedback and flight direction on situational awareness.
\end{blueadd}

\subsection{Results}

\subsubsection{Objective Measurements}

\begin{figure*}[ht]
    \includegraphics[width=\textwidth]{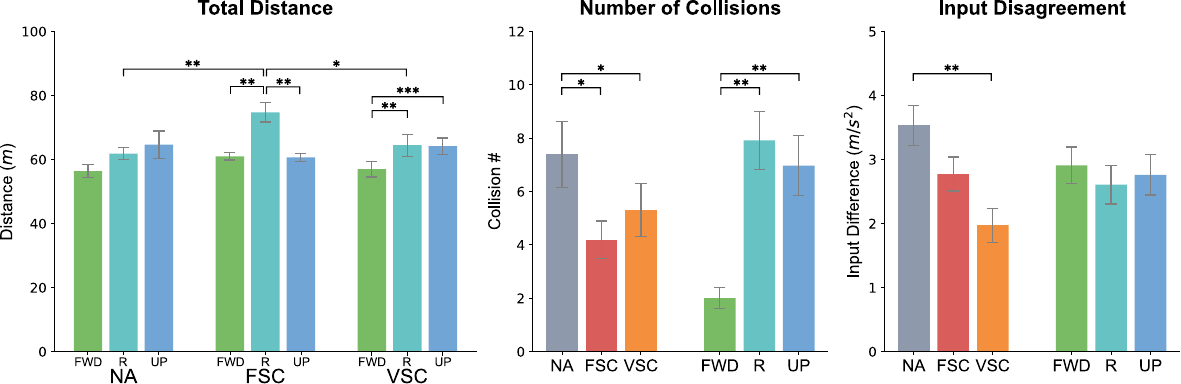}
    \caption{The objective results grouped by feedback condition (NA - no feedback, FSC - force shared control, VSC - vibrotactile shared control) and flying direction (FWD - forward, R - right, UP - upward). \begin{blueadd}The number of collisions and input disagreement for each flying direction were averaged across all feedback conditions.\end{blueadd} The error bars represent the standard error of the mean (SEM), $\oneS p < 0.05$, $\twoS p < 0.01$, and $\threeS p < 0.001$.}
    \label{fig:sec5_performance_metrics}
    \vspace{-10pt}
\end{figure*}

\begin{blueadd}

For total distance travelled, a two-way RM-ANOVA indicated significant main effects of feedback condition \((F(2,22)=3.585,\, p<.05)\) and flying direction \((F(2,22)=17.868,\, p<.001)\), and a significant interaction \((F(1.957,21.527)=4.916,\, p<.01)\). Post-hoc tests showed that FSC-R covered more distance than FSC-FWD, FSC-UP, NA-R \((p<.01)\), and VSC-R \((p<.05)\). VSC-FWD covered less distance than VSC-R \((p<.01)\) and VSC-UP \((p<.001)\).

For the number of collisions, both feedback condition \((F(2,22)=8.095,\, p<.01)\) and flying direction \((F(2,22)=15.653,\, p<.001)\) had significant effects, but there was no interaction \((p = 0.168\)). Post-hoc comparisons showed that NA led to more collisions than FSC and VSC \((p<.05)\), and flying forward yielded fewer collisions than flying right or upward \((p<.01)\).

For input disagreement, feedback condition was significant \((F(2,22)=4.798,\, p<.05)\), but neither flying direction \((p = 0.476\)) nor the interaction was significant \((p = 0.200\)). Post-hoc tests found VSC elicited less disagreement than NA \((p<.01)\).

In summary, VSC resulted in fewer collisions and lower input disagreement compared to NA, and flying to the right or upward increased collisions relative to forward. Thus, \textbf{H1} and \textbf{H2} were accepted.

\end{blueadd}

\subsubsection{Subjective Measurements}


Friedman tests determined that feedback condition had a significant effect on physical demand ($\chi^2(2)=12.950, p<0.01$), effort ($\chi^2(2)=7.294, p<0.05$), overall task load ($\chi^2(2)=6.048, p<0.05$), Q3 Matching Intention ($\chi^2(2)=7.056, p<0.05$), and Q4 Haptic Usefulness ($\chi^2(2)=21.378, p<0.001$). Post-hoc pairwise comparisons revealed that NA required less physical demand than FSC ($p<0.01$), VSC required less effort than FSC ($p<0.05$), and VSC was more helpful than FSC ($p<0.05$). Since significant differences were only found for some metrics, \textbf{H3} and \textbf{H4} were not accepted.


\begin{figure*}[!t]
    \centering
    \includegraphics[width=\textwidth]{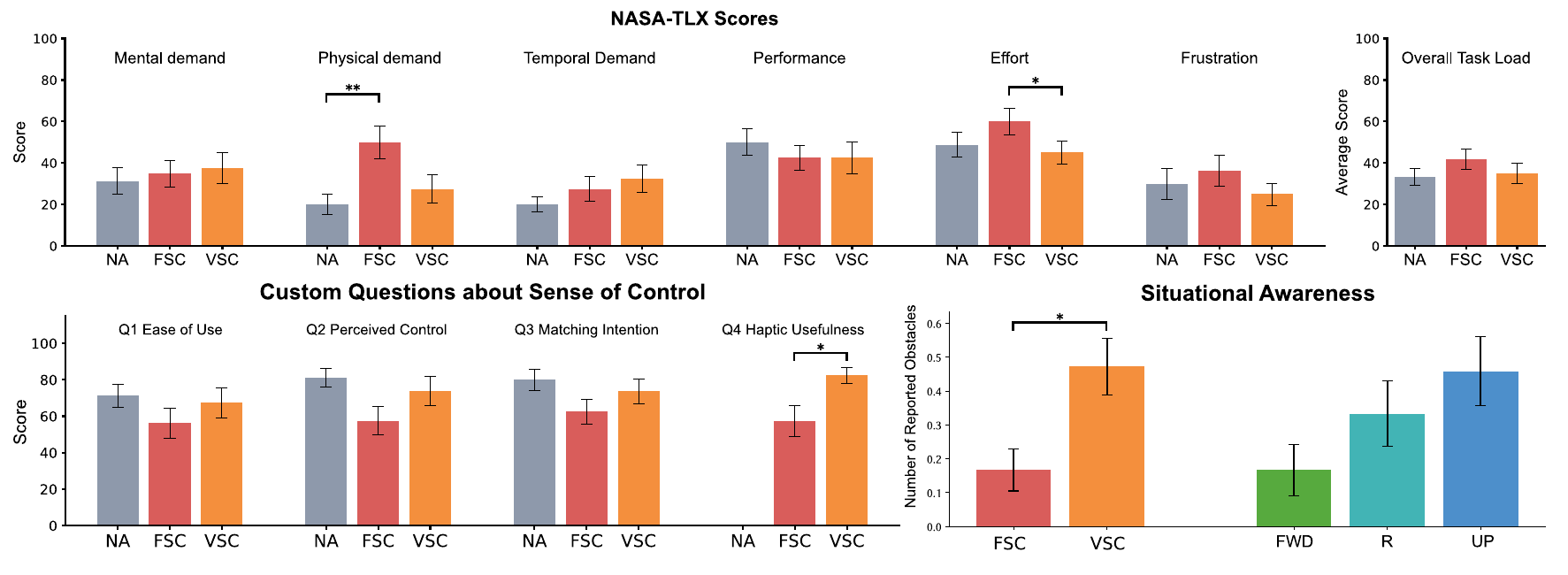}
    \caption{The subjective measurements and situational awareness results grouped by feedback condition (NA - no feedback, FSC - force shared control, VSC - vibrotactile shared control) and flying direction (FWD - forward, R - right, UP - upward). \begin{blueadd}The situational awareness results for each flying direction were averaged across FSC and VSC.\end{blueadd} The error bars represent the standard error of the mean (SEM), $\oneS p < 0.05$, $\twoS p < 0.01$, and $\threeS p < 0.001$.}
    \label{fig:sec5_questionnaire}
    \vspace{-10pt}
\end{figure*}


\subsubsection{Situational Awareness}

\reddelete{
The numbers of reported haptic obstacles are reported in Figure \ref{fig:sec5_questionnaire}. the RM-ANOVA determined that feedback condition had a significant effect ($F(1,11) = 7.436, p < 0.05$), while flying direction did not ($F(2,22) = 2.131, p = 0.143$). The interaction between feedback condition and flying direction was also not significant ($F(2,22) = 0.517, p = 0.604$). Post-hoc pairwise comparisons revealed that VSC was more effective at enhancing situational awareness than FSC  ($p < 0.05$). Therefore, \textbf{H5} was accepted.
}

\begin{blueadd}
An RM-ANOVA revealed a significant effect of feedback condition \((F(1,11)=7.436, p<0.05)\) on situational awareness but no effect of flying direction \((F(2,22)=2.131, p=0.143)\) or an interaction \((F(2,22)=0.517, p=0.604)\). Post-hoc tests found that VSC was more effective for situational awareness than FSC \((p<0.05)\), so \textbf{H5} was accepted (Figure~\ref{fig:sec5_questionnaire}).
\end{blueadd}

\section{Discussion}

In this section, we discuss the design implications that arose from the experimental results.

\subsection{Benefits of Vibrotactile Feedback}

Vibrotactile feedback solved the direct coupling problem inherent in haptic joystick systems and ensured stable operation. When using FSC, operators struggled to understand the intent of the force feedback and often tried to override the input by using manual force in the opposite direction. This resulted in risky and oscillatory movements in the tunnel, which was demonstrated by the longer flying distances and higher input disagreement. When using VSC, however, the decoupling of the input and output channels via the RC controller and AeroHaptix provided operators with full control over the UAV. Operators had sufficient time to process the vibrotactile cues and then steer the UAV towards a safer trajectory. Thus, VSC promoted safer and more cautious operation, reducing the risk of collisions.

Additionally, vibrotactile feedback improved situational awareness by delivering multidirectional cues via the MultiCBF algorithm (i.e., more obstacles were identified through haptic feedback when using VSC than FSC). Participants could effectively utilize vibrotactile feedback and became more aware of the environment.

Although the subjective assessment of VSC (i.e., task workload and sense of control) was not found to be statistically different from FSC, this may be due to the short duration of the experimental tasks, which lasted approximately 1 minute per condition. This prevented participants from experiencing the types of workloads typically associated with longer, real-world teleoperation. Future research should evaluate long task durations to gain a deeper understanding.

\subsection{Effect of Visual Capacity}


The isolated flying directions helped determine how visual capacity influenced participants' reliance on haptic feedback. When obstacles were out of sight in right and upward directions, there were three times more collisions than the forward direction. This suggested that haptic feedback was more helpful in low visual capacity conditions. Due to the limited obstacles in the tunnels, no statistical differences were found between the flying directions for situational awareness. Future research should increase the number of obstacles to determine the effectiveness of haptic feedback in low visibility scenarios.

In addition to flying directions, factors such as light sources and dust \cite{hung2024uav} can also affect visual capacity. Future research should investigate these factors and refine algorithms to dynamically adjust haptic feedback.

\subsection{Intuitive Mappings from Layout Optimization}

Our layout optimization process eliminated perceptual biases and ensured an intuitive mapping between body positions and obstacle directions, which helped reduce perceived workloads. As no significant differences were found between the NA and VSC conditions, the addition of on-body vibrotactile feedback did not impose extra workloads.

FSC, however, led to significantly higher physical demand than NA and required more effort than VSC. This suggests that the removal of the RC controller and the direct coupling of input and output channels imposed potential control challenges. Additionally, previous approaches that rendered force feedback often overlooked the anisotropy of force perception on the hand~\cite{van2013anisotropy,tan2006force}. These perceptual differences could lead to extra effort. Thus, these findings reinforce the importance of addressing perceptual biases in future haptic device design for UAV teleoperation.

\section{Limitations and Future Work}

\reddelete{Although the results showed that on-body vibrotactile feedback was effective at enhancing collision avoidance during UAV teleoperation, there are a few limitations.}\begin{blueadd}There are a few limitations with the system implementation and evaluation design.\end{blueadd} First, upper-body vibrations are limited in representing obstacles below the drone. Currently, we mapped them to the lower back actuators. In the future, We plan to extend vibrotactile feedback to the upper limbs or lower body to alleviate this problem. \begin{blueadd}Second, the real-life setup relied on visual alignment, introducing potential human error. Attaching retroreflective markers to actuators and using motion capture systems could improve alignment precision. Third, the layout optimization produced a universal layout rather than personalized ones. While practical and fair for evaluation, it may overlook individual perceptual differences. Future work should explore efficient methods for creating personalized layouts to enhance performance.\end{blueadd}


Additionally, UAV teleoperation was simulated in a simplified virtual environment, without considering real-world factors such as communication delays, inaccurate UAV state estimations, or control input constraints \cite{lam2007collision}. To understand the role of such factors, we plan to integrate AeroHaptix into commercial UAVs \reddelete{using developer tools such as DJI SDKs.} \begin{blueadd}for real-world experiments. For example, as DJI Onboard SDKs provide access to a drone's sensor data, we can adapt our MultiCBF implementation to process obstacle data from DJI APIs and generate corresponding collision risk responses.\end{blueadd}

Finally, we only modulated the actuator positions and vibration intensities to convey obstacle directions. Future work could utilize more vibration parameters to enrich the information provided about obstacles, such as obstacle types and mobility. As prior research has suggested, the use of frequency, spatial, and temporal patterns could also be viable when rendering such information \cite{de2017designing}. 

\section{Conclusion}


This letter introduces AeroHaptix, a novel vibrotactile feedback system for collision avoidance during UAV teleoperation. The system comprises custom hardware with high-density actuators, fine-grained control, and low-latency communication. An optimal actuator layout iwas derived from a perceptual study to ensure the uniform coverage of obstacle directions. Using a novel multi-point feedback algorithm \emph{MultiCBF}, AeroHaptix was found to reduce collisions and input disagreement and improve situational awareness without inducing extra workload. We encourage future researchers to apply and extend our research to further innovate the design of haptic systems to improve UAV teleoperation.



\bibliographystyle{IEEEtran}
\bibliography{main}



\vfill

\end{document}